\def\year{2019}\relax
\documentclass[letterpaper]{article} %DO NOT CHANGE THIS
\usepackage{aaai19}  %Required
\usepackage{times}  %Required
\usepackage{helvet}  %Required
\usepackage{courier}  %Required
\usepackage{url}  %Required
\usepackage{graphicx}  %Required
\usepackage{amsfonts}
\usepackage{amsmath}
\usepackage{amssymb}
\usepackage{mathptmx}
\usepackage{nicefrac}
\usepackage{booktabs} % Pretty tables
\usepackage{makecell} % cell line breaks..
\usepackage{graphicx}
\usepackage[font=small]{subcaption}
\usepackage[font=small]{caption}
\usepackage{xargs} % commandx 
\usepackage{multicol}

\usepackage{color}
\usepackage[textsize=scriptsize]{todonotes}
\definecolor{blue}{RGB}{0, 93, 170}			%Go Big Blue!
\definecolor{darkgreen}{RGB}{0, 102, 0}

\newcommand{\ignore}[1]{}
\newcommand{\typewriter}[1]{\texttt{#1}}
\usepackage{xspace}
\newcommand{\nlisys}{\textbf{\texttt{ConSeqNet}}\xspace}

\usepackage[font=small]{subcaption}
\usepackage[font=small]{caption}

\newcommand{\citet}[1]{\citeauthor{#1}~\shortcite{#1}}
\newcommand{\citep}{\cite}
\nocopyright

\begin{document}
\title{Improving Natural Language Inference Using External Knowledge\\ in the Science Questions Domain}
%\author{Paper 2498} 
\author{
Xiaoyan Wang$^\S$, Pavan Kapanipathi$^\dagger$, Ryan Musa$^\dagger$, Mo Yu$^\dagger$, \vspace{0.5em} \\ \vspace{0.5em}
{\Large \bf Kartik Talamadupula$^\dagger$, Ibrahim Abdelaziz$^\dagger$, Maria Chang$^\dagger$,  }\\
{\Large \bf Achille Fokoue$^\dagger$, Bassem Makni$^\dagger$, Nicholas Mattei$^\dagger$, Michael Witbrock$^\dagger$}
\AND
{\normalsize \rm $^\S$ University of Illinois at Urbana-Champaign }\\
Department of Computer Science\\
Urbana-Champaign, IL, USA \\
\small{xiaoyan5@illinois.edu}
\And
{\normalsize\rm $^\dagger$IBM Research} \\
IBM T.J. Watson Research Center \\
Yorktown Heights, NY, USA \\
\small{\{ramusa, kapanipa, yum, krtalamad, achille, witbrock\}@us.ibm.com} \\ \small{\{ibrahim.abdelaziz1, maria.chang, bassem.makni, n.mattei\}@ibm.com}
}

\maketitle
\begin{abstract}
% Motivation 
Natural Language Inference (NLI) is fundamental to many Natural Language Processing (NLP) applications including semantic search and  question answering. The NLI problem has gained significant attention due to the release of large scale, challenging datasets.
% Why 
Present approaches to the problem largely focus on learning-based methods that use only textual information in order to classify whether a given premise entails, contradicts, or is neutral with respect to a given hypothesis. 
Surprisingly, the use of methods based on  structured knowledge -- a central topic in artificial intelligence -- has not received much attention vis-a-vis the NLI problem. 
% Existing methods for solving the NLI problem are particularly suited to formal reasoning tasks \nick{when?} applying available knowledge bases.  
While there are many open knowledge bases that contain various types of reasoning information, their use for NLI has not been well explored.
% How
To address this, we present a combination of techniques that harness external knowledge to improve performance on the NLI problem in the science questions domain. We present the results of applying our techniques on text, graph, and text-and-graph based models; and discuss the implications of using external knowledge to solve the NLI problem. Our model achieves close to state-of-the-art performance for NLI on the SciTail science questions dataset.

% We present the behavior of our methods to compare the difference between using just the text of the premise and the hypothesis, and the graphical information as well. 

% With this, we explore the possibilities of the use of multiple knowledge sources for existing NLI datasets. 
% Evaluation results
\end{abstract}

\section{Introduction}
\label{sec:intro}

% NLI and Applications
% Natural Language Inference is a fundamental task for natural language understanding~\cite{maccartney2009}. Addressing NLI has shown to improve the performance of tasks requiring natural language understanding, such as semantic search, question answering, and text summarization. The goal of NLI is to determine whether a natural language hypothesis h can be inferred from natural language premise p. Specifically, this is tasked as a classification problem, where give two sentences, hypothesis and premise, the model classifies them based on "entailment", "contradiction", or "neutral".  In recent times, to learn supervised classifiers for NLI multiple large scale datasets have been released~\cite{bowman2015large,KhSaCl17,adina2018multinli}, and hence this task has gained significant attention.~\cite{yin2018end,parikh2016decomposable,KhSaCl17,chen2018}. 

%% Below changes made by KRT

%% Need to add examples too

Natural Language Inference (NLI) -- also known as textual entailment -- is a fundamental task in Natural Language Understanding (NLU)~\cite{maccartney2009}. Progress on the NLI problem has been shown to improve performance on important tasks that require NLU, including semantic search, question answering, and text summarization. In accordance with this, multiple large-scale datasets for NLI have been released~\cite{bowman2015large,KhSaCl17,adina2018multinli}, and the task has garnered significant attention from the NLP community~\cite{parikh2016decomposable,zhao2016textual,KhSaCl17,yin2018end,chen2018}. 

The main goal in the NLI problem is to determine whether a given natural language {\em hypothesis \typewriter{h}} can be inferred from a natural language {\em premise \typewriter{p}}. NLI is often cast as a classification problem: given two sentences -- hypothesis and premise -- the problem lies in classifying the relationship between them into one of three classes: `entailment', `contradiction', or `neutral'. %In addition, the use of entailment models has become a popular way to judge evidence for open domain question answering \cite{BoPaMiYu18}. %and has been specifically identified as a way to increase the accuracy of question answering systems \cite{BoPaMiYu18}.  
In this paper, we restrict our focus in solving the NLI problem to the ``entailment'' class, as it is the most salient to down-stream tasks such as question answering~\cite{KhSaCl17}; we group the other two options into the ``neutral'' class. Specifically, we develop a framework that accurately assesses whether a given premise entails a given hypothesis using background knowledge provided by external sources including WordNet~\cite{wordnet}, ConceptNet~\cite{speer2017conceptnet}, and DBPedia~\cite{auer2007dbpedia}.

% We group the other two relations into the ``neutral'' class. 
% The intuition behind this directed focus is that 

%% NEED A PARA ON WHY THIS IS AN IMPORTANT AI PROBLEM - connect thru NLP 

% Inference can play an important role in the NLI problem

% \kartik{Working on introduction right now}

Usable inference and reasoning methods have been a central contribution of artificial intelligence research. Specifically, reasoning methods and the knowledge bases that support them have played an important role in addressing formal reasoning tasks. Our own previous work~\cite{BoPaMiYu18} associates the QA task with the reasoning and knowledge types required to answer various types of science exam questions. While the introduction of large datasets for NLI has led to it being cast as a classification problem, the potential of pre-existing knowledge has been only minimally explored~\cite{chen2018,marelli2014semeval}.  This knowledge may manifest itself in various forms: as facts consisting of entities and their relationships; as lexical knowledge about the synonyms of entities; or as common-sense concepts and relationships between them.  Exploiting additional knowledge from different knowledge graphs
%, as in the present work, or from more sophisticated knowledge bases, 
may well aid attempts to solve the NLI problem.

\begin{figure}[h]
  \centering
  \vspace{-2mm}
    \includegraphics[width=0.5\textwidth]{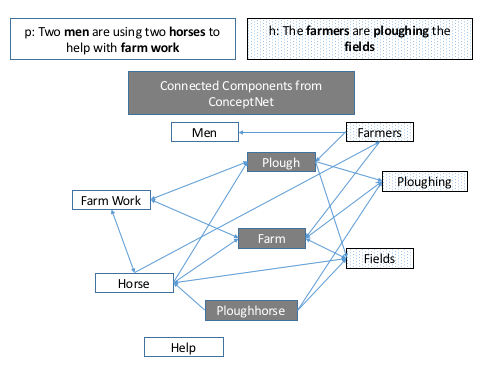}
  \caption{Example of a subgraph from ConceptNet for a given premise \typewriter{p} and a hypothesis \typewriter{h}. Edges represent the existence of a relationship between concepts; nodes (concepts) with a gray background are those not mentioned explicitly in either \typewriter{p} or \typewriter{h}.}
% * <witbrock@us.ibm.com> 2018-09-03T14:09:45.225Z:
% 
% > ConceptNet
% Need a cite here (and the CamelCase spelling is correct)
% Liu H and Singh P: 'Commonsense reasoning in and over natural language', Proceedings of the 8th International Conference on Knowledge-Based Intelligent Information and Engineering Systems (KES-2004) (2004)
% 
% ^.
  \vspace{-2mm}
  \label{fig:ConceptNet_example}
\end{figure}

% Inference has been a key challenge for artificial intelligence and Knowledge bases have played an important role to address this challenge overtime, particularly for formal reasoning tasks. While the introduction of large datasets for natural language inference have enabled research in applying learning algorithms, the potential of knowledge bases have been minimally explored for this task~\cite{chen2018,marelli2014semeval}. In this work, we explore the possibilities and impact of harnessing knowledge bases, in particular knowledge graphs, for natural language inference task. 

%% KRT: Need a couple example here from the csv

%Knowledge bases
% Knowledge bases are extensively used

Figure~\ref{fig:ConceptNet_example} shows an example of a subgraph that can be derived from ConceptNet based on the concepts mentioned in a given premise and hypothesis\footnote{Example based on the SNLI dataset~\cite{bowman2015large}.}. This subgraph includes connections between the concepts mentioned in the premise and the hypothesis, via concepts available in ConceptNet. Such subgraphs, enhances the concepts mentioned in the text with additional information and hence can improve the performance of learning-based systems~\cite{kapanipathi2014user,bouchoucha2013diversified}, particularly for NLI. 

% In this paper, we present a framework that provides the capability to use various kinds of knowledge bases and techniques that can be used to harness relevant knowledge for NLI. Continue the contributions bla bla. 

\smallskip
\noindent
\textbf{Contributions.} In this paper, we introduce the \nlisys framework, which enables the use of various kinds of external knowledge bases to retrieve knowledge relevant to a given NLI instance, by retrieving information related to the premise and hypothesis. We describe our novel architecture and demonstrate its use with a specific external knowledge source -- ConceptNet -- and evaluate its performance on two other sources, WordNet and DBpedia. We compare the performance of three distinct approaches to augmenting the knowledge used to train for and to predict entailment relationships between given pairs of premises and hypotheses: graph-only, text-only, and text-and-graph. Using both qualitative and quantitative results, we demonstrate that introducing graph-based features boosts performance on the NLI problem, but only when text features are present as well. Our system has a competitive performance (accuracy) of $85.2\%$ on the SciTail entailment dataset, which is derived from science domain question answering datasets. 

\section{Background and Related Work}
\label{sec:relatedwork}

We consider prior work in the areas of both the Natural Language Inference, and Knowledge Graphs\footnote{We use External Knowledge and Knowledge Graphs interchangeably in this paper.}; as well as work at their intersection.

\subsection{Natural Language Inference}
\label{sec:relatednli}

%RTE\footnote{\url{http://aclweb.org/aclwiki/index.php?title=Textual_Entailment_Resource_Pool}} and SICK~\cite{marelli2014semeval} datasets were one of the first datasets create for the NLI task. These datasets were small which made it hard to support learning algorithms for NLI tasks. 
Recently, the NLI task has gained significant attention due to the release of multiple crowd-sourced, large-scale datasets that can be used to train neural network classifiers~\cite{parikh2016decomposable,zhao2016textual,KhSaCl17,chen2018,yin2018end}. In particular, the SNLI~\cite{bowman2015large} dataset has been a major catalyst of research on NLI facilitating learning-based approaches by providing a sufficient number of training examples for data-intensive learning algorithms. The Multi Genre NLI corpus (MultiNLI)~\cite{adina2018multinli} is an effort to address the limitations of SNLI. The dataset focuses on domain adaptation by introducing genre labels for each sentence pair. SciTail~\cite{KhSaCl17}, derived from science domain multiple choice question datasets \cite{WLG17a,CEKS16a}, has the focus on the down-stream task of standardized-test question-answering. 
Neural networks with an encoder-attention-classifier architecture are commonly used for NLI tasks~\cite{bowman2015large,wang2017bilateral,KhSaCl17}. To encode the premise and hypothesis, most methods use RNNs, which are used for many NLP tasks. \citet{bowman2015large} used LSTMs to learn sentence representations for premise and hypothesis separately, concatenating them for classification.~\citet{rocktaschel2015reasoning} introduced a word-by-word attention model that learns conditional encodings of premise and hypothesis for textual entailment. The match-LSTM model~\cite{wang2015learning} extended the model presented by \citet{rocktaschel2015reasoning} to address the limitation of a single vector representation of the premise for learning attention weights. The next step in this direction was the multi-perspective matching mechanism introduced by \citet{wang2017bilateral}. Most of these text-based models differ from each other in their attention mechanisms. While the above-mentioned approaches used word-by-word attention mechanisms, \citet{yin2018end} has explored inter-sentence interactions for textual entailment on the SciTail dataset; however, other NLI datasets were not used for evaluation. Recently,~\citet{glockner2018breaking} exposed the simplicity of the existing neural network models by showing that they do not perform well on a new test-set that has significantly lesser overlap between the premise and hypothesis. The best performing model on this test set is~\citet{chen2018} which harnesses WordNet as external knowledge; providing added motivation to explore external knowledge sources for NLI.  

\subsection{External Knowledge (Knowledge Graphs)}
\label{sec:relatedkgs}
Most openly available external knowledge sources are in the form of graphs, commonly known as Knowledge Graphs (KG). Some of the well known KGs include Freebase, DBpedia, Yago, and ConceptNet. A KG is defined as a set of concepts connected by relationships where the concepts form the nodes of the graph and the relationships are the labeled edges. A fact such as  ``Barack Obama is the spouse of Michelle Obama'' is represented in a KG (for example on DBpedia) as \texttt{dbr:Barack\_Obama} \texttt{dbo:spouse} \texttt{dbr:Michelle\_Obama}, where \texttt{dbr:Barack\_Obama} and \texttt{dbr:Michelle\_Obama} are nodes and \texttt{dbo:spouse} is a labeled edge. 

%Such facts in knowledge graphs can be leveraged to augment text by expanding the concepts in the text to include related concepts from the knowledge graph. Techniques that augment text with concepts from knowledge graphs have already been shown to be beneficial~\cite{hoffart2012kore,kapanipathi2014user}\pavan{(add relevant citations)}. Inspired by such works, this paper explores harnessing knowledge graphs for NLI. 

KGs have been used extensively in Information Retrieval~\cite{arbi2014}, Recommendation Systems~\cite{lalithsena2017domain,kapanipathi2014user} and Question Answering~\cite{sun2018open}. 
% and Natural Langugage Processing~\cite{hoffart2012kore}.
%Knowledge bases \textendash{} specifically knowledge graphs such as DBpedia~\cite{auer2007dbpedia} and Yago~\cite{suchanek2007yago} \textendash{} have been shown to improve performance on tasks ranging from query expansion and movie recommendations to entity recognition and disambiguation. 
The prominent concern when using KGs is the availability of relevant knowledge, both in terms of applicability to the task setting and to the domain/topic of interest. This is closely related to the way in which these knowledge graphs are created. For instance, DBpedia is a generic knowledge base with information extracted from Wikipedia infoboxes; these contain facts such as (\textit{Barack\_Obama, spouseof, Michelle\_Obama}). By contrast, ConceptNet~\cite{liu2004ConceptNet} consists of common-sense knowledge acquired through crowd sourcing. While DBpedia is well suited for entity-based tasks such as movie recommendation and entity disambiguation, ConceptNet may be more appropriate if common sense reasoning is required. WordNet is a lexical database offering  synonyms, hypernyms, hyponyms, and antonyms. In this work, we experiment with DBpedia, ConceptNet, and WordNet.

%\mjw{this paragraph is partially redundant with content a couple of paragraphs above} Knowledge graphs created for a variety of purposes may provide differentiated sources of knowledge for NLI. Knowledge graphs such as DBpedia, Yago, and Freebase are primarily concerned with representing factual relationships between named entities.  Other knowledge bases such as ConceptNet and Cyc have been constructed to be used for common sense reasoning, and contain relationships such as relatedTo, usedFor, primaryConstituent, and partTypes. 

\subsection{Knowledge Graphs and NLI}
\label{sec:realtednlikg}
% KB Based NLI Approaches
Few prior approaches have exploited syntactic structures in the form of graphs for textual entailment~\cite{zhao2016textual,KhSaCl17}. 
% The Decomposed Graph Entailment Model~\cite{KhSaCl17} (DGEM) -- which is an extension to the Decomposable Attention model~\cite{parikh2016decomposable} (DecompAtt) --  uses OpenIE to construct syntactic structures (as graphs) for hypothesis, and computes node attention based on the premise text. 
However, the graph structures used in these models do not come from external knowledge sources and are not used to enhance the textual content of either the premise or hypothesis. One work which closely relates to the objective of the paper, i.e., enhancing the model with external knowledge is~\citet{chen2018}. This work harnesses WordNet as the external knowledge for NLI. WordNet, however, is a lexical database restricted to a small number of linguistic relationships among terms. Furthermore, \citet{chen2018} uses only four relationships to generate five features based on WordNet. In our work, including WordNet, we explore more expressive KGs such as DBpedia and ConceptNet. Also, the most important difference is the way we use KGs: our method enriches the texts (as graphs) and builds a matching model over the enriched texts for textual entailment. Another interesting alternative direction for using KGs such as WordNet for NLI is presented by \citet{kang2018adventure}.  The approach generates adversarial examples for robust training of NLI systems, but this improves performance only during limited supervision. %The evaluation shows an improvement on existing NLI systems with limited supervision (lack of training data). However, the models are not comparable to any state of the art NLI systems in its performance.

%\subsection{Knowledge Graph Embeddings}

%Several graph embedding techniques~\cite{transe,transh,hole,complex} have been proposed recently to embed an input graph into a continuous vector space, and use low-dimensional embedding vectors to represent entities and relationships. TransE~\cite{transe} is an early proposed approach which models a relation ($r$) as a translation from a head entity ($h$) to a tail entity ($t$). It tries to align the subject vector as closely as possible to the object vector once translated by the relation vector. To learn these embeddings, TransE uses a pair-wise ranking loss function which measures the score of each possible triple as the distance between $h + r$ and $t$. TransH~\cite{transh} improves over TransE by allowing entities to have different embeddings according to the relation. To do so, the head and tail entities are first projected to the relation hyperplane before computing the distance between them. Instead of simple addition of vectors, HolE~\cite{hole} uses more expressive operators to model a triple. It captures richer interactions by using the circular correlation of vectors to represent pairs of entities. Similarly, ComplEX~\cite{complex} represents entities and relations with complex-valued embeddings, where a triple is scored using the real part of the multi-linear dot product of the corresponding embeddings. 

\section{Approach}
\label{sec:approach}

\begin{figure*}[h]
  \centering
    \includegraphics[width=0.8\textwidth]{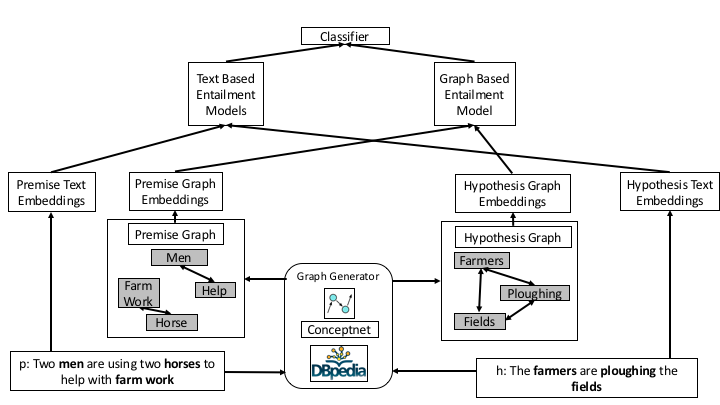}
  \caption{The overall architecture of the \nlisys~system, illustrated via an example.}
  \label{fig:arch}
\end{figure*}

In this section, we present the architecture of our system called \nlisys, and the approach underlying it. Figure~\ref{fig:arch} depicts the framework of \nlisys\ --  the system can use both textual as well as structured information from KGs to assist in determining textual entailment. The framework can be clearly divided into two parts: (a) a {\bf text based model} that takes in as input the premise and hypothesis text; and (2) a {\bf graph based model} whose input is specific knowledge derived from the knowledge base using the given premise and hypothesis. 
% We present the details of these models in turn.
%The framework has the flexibility to either jointly or individually train the above mentioned models. 
 
\subsection{Text Based Model}
\label{sec:textbasedmodel}

Most models developed for NLI have relied on the text of the given premise and hypothesis, without the aid of any external knowledge. These models follow an encode-attend-classify framework. First, the premise and hypothesis are encoded using recurrent neural networks (RNNs). Next, an attention layer is implemented on top of the encoders. The final layer is then used for classification. In our work, we have primarily used match-LSTM~\cite{wang2015learning} as our text based model. We opted for match-LSTM for two main reasons: (1) match-LSTM as an entailment model has been proved useful for multiple NLP tasks~\cite{wang2015learning,wang2017evidence} 
% has been a critical component of multiple reading comprehension techniques; 
and (2) our implementation of match-LSTM performed significantly better than the baselines (details in Section~\ref{sec:evaluation}). 
% While the DecompAttn model has been adapted previously by DGEM, to the best of our knowledge there is no prior work that adapts  match-LSTM as an entailment model on the SciTail dataset. 

%has not been used for the SciTail dataset before our testing. Our implementation of match-LSTM shows an improvement in performance in comparison to other entailment models. Hence, in this work, we use a modified version of match-LSTM on the SciTail dataset.

%\begin{figure}[h!]
%\begin{center}
%\begin{tabular}{lcc}  
%\toprule
%Model & Dev	& Test \\
%\midrule 
%DecompAttn~\cite{parikh2016decomposable}	& 75.4	& 72.3 \\
%DGEM~\cite{wang2015learning} & 79.6 & 77.3 \\
%CAFE & -- & 83.3 \\
%DeIsTe~\cite{yin2018end} & 82.4 & 82.1 \\
%match-LSTM~\cite{wang2015learning}	& 88.2	& 84.1 \\
%\bottomrule
%\end{tabular}
%\end{center}
%\caption{Performance of entailment models on SciTail.}
%\label{tab:entailment_performance}
%\end{figure}

% \subsubsection{match-LSTM}
% Match-LSTM primarily has two components: the attention layer and the matcher. 
Given a premise $P$ (with $K$ words) and a hypothesis $H$ (with $J$ words), the model computes the matching results between them as follows:

\begin{itemize}
\item \textbf{Context Encoding:}
A contextual representation of the premise and hypothesis is generated by first transforming premise $P$ and hypothesis $H$ into their embedding vectors $\mathbf{t^p_i}$ and $\mathbf{t^h_j}$, where $\mathbf{t^p_i}$ and $\mathbf{t^h_j}$ are embedding vectors of $i$-th word in premise and $j$-th word in hypothesis. These embedding vectors are further encoded using BiLSTMs to generate the context encodings of premise and hypothesis. Let $\mathbf{p}_i$ and $\mathbf{h}_j$ be the contextual representation of the $i$-th word of premise and the $j$-th word of hypothesis.

\item \textbf{Word-by-Word Attention:} This layer computes the inter-attention between the contextual embeddings of the premise and hypothesis. The entries of the (unnormalized) attention matrix $\mathbf{E}\in\mathbb{R}^{K\times J}$ are defined as:
\begin{equation}
E_{ij} = \mathbf{p}_i\cdot\mathbf{h}_j
\end{equation}
we can then compute the soft alignment $\alpha_j$ for the hypothesis as follows:
\begin{equation}
 \alpha_j = \sum_{i=1}^{K}\frac{\exp(E_{ij})}{\sum_{k=1}^{K}\exp(E_{kj})}\mathbf{p}_i
\end{equation}

\item \textbf{Matcher:} We compare the soft aligned premise and hypothesis at each word position as a feature vector:
% \begin{equation}
% \tilde{\mathbf{p}}_i = F\left(\left[\mathbf{p}_i; \beta_i; \mathbf{p}_i - \beta_i; \mathbf{p}_i \odot \beta_i\right]\right)
% \end{equation}
\begin{equation}
\tilde{\mathbf{h}}_j = \left[\mathbf{h}_j; \alpha_j; \mathbf{h}_j - \alpha_j; \mathbf{h}_j \odot \alpha_j\right], \nonumber
\end{equation}
where $\odot$ denotes the element-wise multiplication, and $[;]$ denotes vector concatenation.
The feature vectors for all the hypothesis positions are fed into a BiLSTM to get the matching states $\{\mathbf{h}_j^\textrm{m}\}_{j=1:J}=\textrm{BiLSTM}(\{\tilde{\mathbf{h}}_j\}_{j=1:J})$.

\item \textbf{Pooling:} To get a fixed-sized vector representation for the matching result, we apply max-pooling:
\begin{equation}
% \mathbf{x}_{\text{out}}^{\text{text}}=\mathbf{h}^\mathrm{m}_{\max} = \max\left(\left[\mathbf{h}_1^\textrm{m}, \mathbf{h}_2^\textrm{m}, \dots, \mathbf{h}_N^\textrm{m}\right]\right)
\mathbf{x}_{\text{out}}^{\text{text}} = \max\left(\left[\mathbf{h}_1^\textrm{m}, \mathbf{h}_2^\textrm{m}, \dots, \mathbf{h}_J^\textrm{m}\right]\right)
\end{equation}
$\mathbf{x}_{\text{out}}^{\text{text}}$ is then used for classification. 
\end{itemize}

% Then, an attention mechanism is used to determine the attention weighted representation of a premise as follows:

% \begin{multicols}{2}
% \begin{equation}
% a_k = \sum_{r=1}^{K}\alpha_{kr} h_r^p
% \end{equation}\break
% \begin{equation}
% \alpha_{kr} = \dfrac{\exp(e_{kr})}{\sum_{r'}\exp(e_{kr'})}
% \end{equation}
% \end{multicols}

% \noindent where
% \begin{equation}
% e_{kr} = w_e \mathbf{.} \tanh(W^p h_r^p + W^h h_k^h + W^m h_{k - 1}^m)
% \end{equation}

% \noindent In Equation 3, $W^i$ are the weights to be learned. $h_{k - 1}^m$ is the $k^th$ hidden state of the generated by the matcher. Matcher is $LSTM(m)$ where $m_k = [a_k;h_k^h]$. 
% The max-pooling result over the hidden states $\{\mathbf{h_j^m}\}_{j=1:N}$ of the matcher is used for classification. 

%. Difference with Neural Attention Model -- Mathcher~\ibrahim{do we need to rephrase this sentence or add a reference?} concatenates the attention-weighted representation of premise with hypothesis to produce the final result. 

\subsection{Graph Based Model}
\label{sec:graphbasedmodel}

The input to the graph based model is a premise graph and a hypothesis graph respectively. As shown in Figure~\ref{fig:arch}, the first step is to transform the given premise and hypothesis text into a relevant subgraph mapped to an external knowledge. Given premise text $P$ and hypothesis text $H$, the transformation generates $P_g = (V_p, E_p)$ and $H_g = (V_h, E_h)$ where $V_p = (v^p_1,...,v^p_K)$ and $V_h = (v^h_1,..., v^h_J)$ are a subset of concepts in the knowledge graph. $E_p$ and $E_h$ are labeled edges connecting concepts in $P_g$ and $H_g$ respectively. $H_g$ and $P_g$ are  the input for our graph-based entailment model. 

\subsubsection{Premise and Hypothesis Graphs}
\label{sec:graphgeneration}

%The transformation to the premise and hypothesis subgraphs $P_g$ and $H_g$ is achieved by mapping the phrases in $P$ and $H$ to concepts in the knowledge graph, respectively. 
Premise and hypothesis graphs are generated by mapping text phrases to concepts in the external knowledge graph.  For instance, as shown in Figure~\ref{fig:ConceptNet_example}, given hypothesis \textit{``The farmers  are ploughing the fields"},  we map ``\textit{farmers}'', ``\textit{ploughing}, and ``\textit{fields}" in the sentence to their corresponding concepts in the external knowledge. %These mappings are used to select the relevant subgraphs $P_g$ and $H_g$, for premise and hypothesis from the external knowledge source. To select the relevant subgraphs we have developed the following methodologies:
We use the following techniques to generate three different types of subgraphs for each premise and each hypothesis separately:

\begin{itemize}
\item \textit{Concepts Only\footnote{Concepts in the \textit{Concepts Only} graph are used to form the base set for \textit{One-Hop} and \textit{Two-Hop} graph generation techniques.}:} %The concepts that are mapped from premise and hypothesis text to the external knowledge form the vertices of the corresponding subgraphs. The edges between them in the external knowledge form the edges in the premise and hypothesis graphs. The premise graph $P_g$ therefore consists of concepts mentioned in the text of premise, and the edges between them from external knowledge. The same approach is followed for the hypothesis.
For each premise and hypothesis, we use lexical mappings within each knowledge source to map individual words and phrases to concepts.  Those concepts make up the vertices of the \textit{Concepts Only} subgraph.  All edges connecting those vertices are also included in the subgraph.  

\item \textit{One-Hop:} %The concepts in the \textit{Concepts-Only} graph for premise and hypothesis are expanded to include all their one hop neighbors in the external knowledge source.
The concepts in the \textit{Concepts Only} graph for premise and hypothesis are expanded to include all their one hop neighbors in the external knowledge source.

\item \textit{Two-Hop:} %On average, the \textit{One-Hop} graphs generated using ConceptNet expand the number of concepts from 9 to 326; this is a large set and may be noisy. To address this issue, we create another graph by selecting paths that connect the concepts already mapped in premise (for premise graph) and in hypothesis (for hypothesis graph). For instance, the premise graph $P$ will include paths of two hop whose source and the destination concepts are mapped from premise text to external knowledge\footnote{These concepts are the same as those in the \textit{Concept-Only} graph of premise}.\pavan{Kartik, is this understandable now?} 
Two-Hop graphs were generated by adding two hop neighbors to the \textit{Concepts Only} graph, but only if they constitute paths between vertices that were already in the \textit{Concepts Only} graph. This can be thought of as a graph that increases interconnectedness by introducing minimal new concepts, hence reducing noise. We took this approach because we observed that \textit{One-Hop} graphs turned to be noisy (on average, the \textit{One-Hop} graphs generated using ConceptNet increased the number of concepts extracted from premise text from 9 to 326).

% \item Inter Sentence: Paths of two hop where, for the premise graph, the source nodes are the concepts in premise concept only graph and the destination nodes are the concepts in the hypothesis concepts only graph. Vice versa for the hypothesis graph.
\end{itemize}
\noindent We use the graphs constructed (under the three conditions above) for premise and hypothesis as the input to our neural network model for entailment. In this work, we explore two different neural network models: (1) Gmatch-LSTM; and (2) Graph concepts attention model (GconAttn).

%\begin{figure}[h]
%  \centering
%   \vspace{-0.1in}
%    \includegraphics[width=0.45\textwidth]{images/graph_model.png}
%     \vspace{-0.2in}
%  \caption{The graph model in \nlisys, which is used to cast external knowledge into the form of a graph.}
%  \label{fig:graph_model}
%\end{figure}

\paragraph{Gmatch-LSTM for Concepts Only Graphs}
Our first design choice was to use match-LSTM with the \emph{Concepts Only} graphs $P_g$ and $H_g$.
The key idea is to treat each node (concept) in the vertex set of a given graph ($V_p$ or $V_h$) as a single token, and re-arrange them into a sequence of concepts. This sequence is ordered by the length of the concept and its positions in the original texts. Thus, match-LSTM can be applied to match the sequence of premise concepts and the sequence of hypothesis concepts.
% match-LSTM~\cite{wang2016machine} on SciTail (Table~\ref{tab:entailment_performance}) performs better than most of the baselines. 
% However, match-LSTM uses a RNN to get a contextualized representation of its input and a RNN in the matcher. RNNs tries to capture the sequence information in text.

In contrast with the text based model, where each word is first represented as its word embedding vector, in the graph based match-LSTM each concept is initialized using the corresponding \textit{Concept Embedding}. The concept embeddings are trained by knowledge graph embedding techniques such as TransH~\cite{transh}, and CompleX~\cite{complex} using the corresponding knowledge graph.

\paragraph{GconAttn Model for General Graphs}

The use of match-LSTM is justified when the input is the \textit{Concepts Only} graph, since the concept can be aligned in the order of their appearance in premise and hypothesis to form text sequences. However, it is non-trivial to order the concepts in the graphs generated by the \textit{One-Hop} and {\it Two-Hop} strategies; this makes the use of match-LSTM unintuitive. We therefore develop a new model {\em GconAttn} (Graph Concepts Attention Model) to overcome this limitation.

% The match-LSTM can only be justified for its use for \textit{Concepts ONLY} graph, because the concepts in the graph can be aligned to the order of its mappings in premise and hypothesis text thus could form text sequences. However, it is non-trivial to order the concepts in graphs generated from \textit{One Hop} and \textit{Two Hop} strategy which makes it unintuitive to use match-LSTM as a graph model. Therefore, we developed a simple, new model, \emph{GDecomp Attention}, to overcome the above mentioned limitations and use it for all the graphs. 

%The neural network model is based on attention is inspired by DecompAttn Model~\cite{parikh2016decomposable} (RECHECK THIS) with a few modifications. 
%The overall schematic of the graph based model is depicted in Figure~\ref{fig:graph_model}. 
The embeddings of premise and hypothesis concepts from the graph are the input to the neural network. An attention weighted representation of both premise and hypothesis is determined, and is transformed to a final fixed size representation using pooling. This representation of premise and hypothesis (concatenated) is used for classification. Below, we distinguish the graph based model from the match-LSTM (text based) model described previously: 
% \begin{itemize}
% \item Given the embeddings for each entity trained using one of the KG embedding techniques on the corresponding knowledge graph, we use attention mechanism to compute similarities between premise and hypothesis. 
% \item Feed forward over hypothesis tilda and premise tilda
% \item Max pooling
% \item Classification
% \item Also discuss modification of the graph to use match-LSTM
% \end{itemize}

% \noindent For simplicity, let $S$ denote either premise $P$ or hypothesis $H$.

\begin{itemize}

\item \textbf{Context Encoding:\;} Since the concept graphs may not have sequential structures, the GconAttn model directly starts with the {\bf concept embeddings}.

\item \textbf{Word-by-Word Attention:\;} This layer computes the inter-attention between the embeddings of the concepts in premise and hypothesis to find the best aligned concepts between the respective graphs. The major difference from match-LSTM is that GconAttn performs two-way attention~\cite{seo2016bidirectional}.
The model computes the soft alignment for premise and hypothesis respectively as follows:
\begin{equation}
\beta_i = \sum_{j=1}^{J}\frac{\exp(E_{ij})}{\sum_{k=1}^{J}\exp(E_{ik})}\mathbf{h}_j, \quad \alpha_j = \sum_{i=1}^{K}\frac{\exp(E_{ij})}{\sum_{k=1}^{K}\exp(E_{kj})}\mathbf{p}_i \nonumber
\end{equation}

\item \textbf{Matcher:\;} Due to the lack of sequential structure, the matching nodes of each hypothesis concept $j$ or premise concept $i$ are computed with a projection feed-forward network:
\begin{equation}
{\mathbf{p^m_i}} = F\left(\left[\mathbf{p}_i; \beta_i; \mathbf{p}_i - \beta_i; \mathbf{p}_i \odot \beta_i\right]\right) \nonumber
\end{equation}
\begin{equation}
{\mathbf{h^m_j}} = F\left(\left[\mathbf{h}_j; \alpha_j; \mathbf{h}_j - \alpha_j; \mathbf{h}_j \odot \alpha_j\right]\right) \nonumber
\end{equation}
where $F(\cdot)$ denotes a feed-forward network; $\odot$ denotes the element-wise multiplication; and $[;]$ denotes vector concatenation.

\item \textbf{Pooling:\;} Finally, we apply max-pooling and mean-pooling across all matching nodes in the premise and hypothesis. The output of pooling is concatenated, and is presented as the output of the graph model, which can then be used to make the final prediction. $\mathbf{s} \in \{\mathbf{h},\mathbf{p}\}$.
\begin{align}
\mathbf{s}'_{\max} = \max\left(\left[{\mathbf{s^m_1}}, {\mathbf{s^m_2}}, \dots, {\mathbf{s^m_N}}\right]\right)\\
\mathbf{s}'_\text{avg} = \text{avg}\left(\left[{\mathbf{s^m_1}}, {\mathbf{s^m_2}}, \dots, {\mathbf{s^m_N}}\right]\right) \nonumber\\
\mathbf{x}_{\text{out}}^{\text{graph}} = \left[\mathbf{p}'_{\max}; \mathbf{p}'_\text{avg}; \mathbf{h}'_{\max}; \mathbf{h}'_\text{avg} \right] \nonumber
\end{align}

% \begin{equation}
% \mathbf{s}'_\text{avg} = \text{avg}\left(\left[{\mathbf{s^m_1}}, {\mathbf{s^m_2}}, \dots, {\mathbf{s^m_N}}\right]\right) \nonumber
% \end{equation}

% \begin{equation}
% \mathbf{x}_{\text{out}}^{\text{graph}} = \left[\mathbf{p}'_{\max}; \mathbf{p}'_\text{avg}; \mathbf{h}'_{\max}; \mathbf{h}'_\text{avg} \right] \nonumber
% \end{equation}
\end{itemize}

\subsection{Merging Text and Graph Models}

The final results of  matching the text model $\mathbf{x}_{\text{out}}^{\text{text}}$ and the graph model $\mathbf{x}_{\text{out}}^{\text{graph}}$ are concatenated, and passed on to a feed forward network that classifies between \textit{entailment} and \textit{neutral}, which are the two classes in the SciTail dataset:

\begin{equation}
\mathbf{pred} = F\left(\left[\mathbf{x}_{\text{out}}^{\text{text}}; \mathbf{x}_{\text{out}}^{\text{graph}}\right]\right)
\end{equation}

% where $\mathbf{h}^\mathbf{m}_{\max}$ is the final hidden state of the text model (match-LSTM) and $\mathbf{x_{out}}$ is the final hidden state of the graph model. 

\noindent It is important to note that the final hidden state of the graph model can be directly fed into a classifier. This can be considered an entailment model that uses only  information from external knowledge. 

% We have used this model (without textual information) to make decisions on selecting the graph construction technique for our approach (c.f. Section~\ref{sec:implementation}). 

%We do not perform any encoding over the graph embeddings where graphs have concepts not mentioned in premise and hypothesis (one hop and two hop). For the graph construction technique where the nodes have only concepts mentioned in premise and hypothesis, we order the entities based on their position in the premise and hypothesis, respectively, and use DecompAttn or match-LSTM over it \nick{to embeedd them?}. This translates \nick{is the same as?} to using RNNs to encode the graph embeddings.   
\section{Experiments and Results}\label{sec:evaluation}

In this section, we detail the experiments that we perform to evaluate our \nlisys~system. We first describe each dataset, followed by the training setup of our best model and implementation details. We then compare that model's performance to numbers from recent entailment models and baselines. In the latter parts of this section, we discuss the selection of knowledge graphs and the performance of graph construction techniques -- specifically, the \textit{Concepts Only, One-Hop,} and \textit{Two-Hop} methods (c.f. Section~\ref{sec:approach}).

\subsection{Datasets}

We use the SciTail dataset~\cite{KhSaCl17}, which is a textual entailment dataset derived from publicly released science domain multiple choice question answering datasets~\cite{WLG17a,CEKS16a}. The dataset contains $27,026$ sentence pairs (premise and hypothesis), with binary labels denoting whether the relationship between each pair is \texttt{entails} or \texttt{neutral}. The hypothesis is created using the question and correct answer from the options; the premise is retrieved from the ARC corpus (\url{data.allenai.org/arc/arc-corpus}. The performance of the entailment models on this dataset is shown in Table~\ref{tab:entailment_performance}; the evaluation metric is accuracy. 
%It has been shown that the improved performance on the SciTail dataset in turn improves the performance on the multiple choice question answering challenge~\cite{KhSaCl17}. 

\begin{table}[h!]
\begin{center}
\resizebox{\linewidth}{!}{
\begin{tabular}{lcc}  
\toprule
Model & Dev	& Test \\
\midrule 
Decomp-Attn~\cite{parikh2016decomposable}	& 75.4	& 72.3 \\
DGEM*~\cite{KhSaCl17} & 79.6 & 77.3 \\
DeIsTe~\cite{yin2018end} & 82.4 & 82.1 \\
BiLSTM-Maxout~\cite{MiClKhSa18} & - & 84.0 \\
match-LSTM~\cite{wang2015learning}	& 88.2	& 84.1 \\
\midrule
\multicolumn{3}{c}{Our implementation} \\
match-LSTM (GRU) & 88.5 & 84.2 \\
match-LSTM+WordNet*~\cite{chen2018} & 88.8 & 84.3 \\ 
match-LSTM+Gmatch-LSTM* (\nlisys) & \textbf{89.6} & \textbf{85.2} \\
\bottomrule
\end{tabular}
}
\end{center}
\caption{Performance of entailment models on SciTail in comparison to our best model that uses match-LSTM as the text and the graph model with \textit{Concepts Only} graph and CN-PPMI embeddings. * indicates the use of external knowledge in the approach.}
\label{tab:entailment_performance}
\end{table}

\subsection{Training Setup}

All words in the text model are initialized by 300D Glove vectors (Glove 840B 300D) (\url{nlp.stanford.edu/projects/glove}), and the concepts that act as the input for the graph model are initialized by 300D ConceptNet PPMI vectors~\cite{speer2017ConceptNet}; these are openly available for ConceptNet. We use the pre-trained embeddings without any fine tuning. We have adapted match-LSTM with GRUs as our text and graph based model. The system is trained by Adagraph with a learning rate of $0.001$, and batch size of $40$. Both the text and graph based models are trained jointly. For the graph model, we use concepts from the \textit{Concepts Only} graph, which is generated using the approach detailed in Section~\ref{sec:graphgeneration}.

\subsection{Implementation}
\label{sec:implementation}

We used the AllenNLP (\url{allennlp.org}) library to implement all the models used in the experiments. While the previous section (Training Setup) specifically focused on our best performing model, here we provide implementation details for all the experiments that were performed. We used a plug-and-play approach where we varied: (a) text models (match-LSTM, DeCompAttn); (b) graph models (Gmatch-LSTM, GconAttn); (c) external knowledge sources (DBpedia, WordNet, and ConceptNet); (d) graph construction techniques (\textit{Concepts Only, One Hop,} and \textit{Two Hop}); and (e) graph embeddings (CN-PPMI, TransH).

In order to map text to concepts, we used DBpedia Spotlight \cite{mendes2011dbpedia} for DBpedia, and Spacy (\url{spacy.io}) to implement a max-substring match for WordNet and ConceptNet. While ConceptNet had openly available embeddings (CC-PPMI), we used TransH embeddings~\cite{transh} generated using OpenKE (\url{openke.thunlp.org}) with default configurations for the other knowledge sources. 

%We use a 300D for Wordnet and ConceptNet embeddings while we could not generate more than 100D embeddings for DBpedia within a reasonable time ($<$ 24 hours). 

\subsection{Baselines and Comparison}
We compare our work to the following baselines: (1) Decomposable Attention Model (Decomp-Att)~\cite{parikh2016decomposable}; (2) Decomposed Graph Entailment Model (DGEM)~\cite{KhSaCl17}, which is the first of the entailment models to use graph structure from OpenIE as external knowledge and show improvements on SciTail; (3) Deep explorations of Inter-sentence interactions for Textual entailment (DeIsTe)~\cite{yin2018end}; (4) BiLSTM-Maxout~\cite{MiClKhSa18}, the latest entailment model that has shown promise on the SciTail dataset\footnote{The details of the BiLSTM-Maxout~\cite{MiClKhSa18} model are not available. We have reported the numbers provided in a pre-print of a publication that is to appear.}; (4) match-LSTM~\cite{wang2015learning}, which has shown good performance on the SNLI dataset; (5) match-LSTM with GRU, which replaces LSTMs with GRUs for its encoding, since GRUs give better empirical results (this is also the match-LSTM model used in our \nlisys\ model); and (5) match-LSTM -- WordNet features~\citet{chen2018}, which uses five features based on synonyms, antonyms, hypernyms, and co-hypernyms from WordNet (external knowledge) in its co-attention mechanism to improve performance on SNLI. We reimplement these five features and add them to baseline (5).

Table~\ref{tab:entailment_performance} shows the performance of our \nlisys\ model in comparison to other entailment models. Two of the five models in Table~\ref{tab:entailment_performance} use some kind of external knowledge (indicated by *). Almost all the match-LSTM variations exhibit accuracy greater than 84\%, which is better than the recent entailment models on which the SciTail dataset has been tested. Similar to the results shown in~\cite{Yin2017ComparativeSO}, using GRUs with match-LSTM slightly improved the accuracy (particularly on the dev set). The accuracy of our jointly trained model \nlisys\ is $89.6\%$ on the dev set and $85.2\%$ on the test set. Later in this paper, we expound further on our model -- this includes choices such as the use of different knowledge sources, graph models, and the graph construction technique. 
% \pavan{Recent EMNLP paper on ARC from AI2 has a Bi-LSTM Max Out model that has an accuracy of 84.0. We need to discuss whether it is 84 or 85.4. I am inclined to mention the model in the above table if it is 84 otherwise a foot note if 85.4 on the test set. In our future work, we are inclined to test Bi-LSTM Max-Out as text and graph model. }

\subsection{Selecting External Knowledge Source}
In this work, we focus on openly available knowledge graphs. Based on the knowledge graphs' availability for use and their distinct properties, we chose DBpedia, ConceptNet, and WordNet. In Table~\ref{tab:kgstats}, we provide details on each of these knowledge graphs. DBpedia is the largest with more than $5$ million entities and $33$ million facts. ConceptNet subsumes WordNet conceptually, and both contain general type information; however, the reliability of WordNet's linguistic features is higher than ConceptNet's. 

% There are different types of knowledge graphs 
\begin{table}
\begin{tabular}{  |l|c|c|c| }
 \hline
 & DBpedia & WordNet & ConceptNet \\
  \hline			
  Entities/Concepts & 5M & 155K & 1.1M \\
  Relationships & 1100 & 16 & 40\\
  Facts & 33M & 117K & 3.15M \\
  \hline  
\end{tabular}
\caption{Comparison between different knowledge sources based on the number of entities, relationships, and facts.}\label{tab:kgstats}
\end{table}

%Wordnet, DBpedia, and ConceptNet are one of the few relevant knowledge sources that can be exploited for NLI on the SciTail Data. 

For the NLI problem, WordNet has only been partially explored. WordNet is a lexical database with a restricted set of relationships between terms.~\citet{chen2018} use WordNet by creating five new knowledge-based features that are added (for co-attention) to their model. The five features are derived from synonym/antonym and hypernym/hyponym relationships in WordNet between terms in the premise and hypothesis. In Table~\ref{tab:entailment_performance}, we show that using the expanded set of features from WordNet -- used in the attention mechanism of match-LSTM -- has an accuracy of $84.3\%$ on the SciTail datset. While WordNet is useful, it is restrictive in terms of its applicability, particularly because of its coverage and the types of relationships available. DBpedia and ConceptNet are larger knowledge sources, as shown in Table~\ref{tab:kgstats}. 
% We intended to explore both these knowledge bases. 

In terms of the number of concepts mapped from text to external knowledge, ConceptNet has more concepts (9) on average in comparison to DBpedia (6). While it is important that the graphs are not noisy, it is also important to have enough information to exploit. In which case, ConceptNet is slightly better with few more concepts per sentence.

The types of relationships between concepts in DBpedia are factual and derived from Wikipedia. Based on qualitative analysis of these relationships, we found that they may not be suitable for NLI datasets, and specifically SciTail. On the other hand, we found that, ConceptNet with its 40 relationships expressing common sense knowledge may be more suitable. In order to quantitatively select the right knowledge source, we determined the  impact of the each knowledge graph on SciTail. We created the \textit{Concepts Only} graph from each of the available external knowledge sources, and evaluated them using our match-LSTM+GconAttn model on the dev set of SciTail. The results of this experiment are shown in Table~\ref{tab:knowledge_bases_performance}; based on these results, we decided to use ConceptNet for all our graph-based experiments (detailed next). 
% \pavan{Kartik -- Please give this a read.}

\begin{table}[h!]
\begin{center}
\begin{tabular}{lc}  
\toprule
Knowledge Sources & Accuracy\\
\midrule 
WordNet  & 87.6 \\
DBpedia  &  87.3 \\
ConceptNet & \textbf{88.6} \\
\bottomrule
\end{tabular}
\caption{Results using match-LSTM+GconAttn with different knowledge sources on SciTail dev set.}
\label{tab:knowledge_bases_performance}
\label{tab-kgs}
\end{center}
\end{table}

%\subsection{Text to Graph}
%\begin{itemize}
%\item Describe DBpedia spotlight and the light weight concept extractor developed for ConceptNet. 
%\item Figure out how to include wordnet here. Wordnet is necessary because of~\cite{chen2018neural}
%\end{itemize}

%\section{Results and Analysis}\label{sec:results}
%The architecture is setup to plug and play with multiple modules that includes: (1) Word embeddings for the text based model, (2) Different text based models and the integration with the graph based models, (3) Graph embeddings for the graph based model, (4) Graph generation techniques for given premise and hypothesis, (5) Knowledge graphs that support the graph models with external knowledge.\moyu{do we need results of (1) mLSTM with additional features from Chen et al., ACL2018; (2) combination of two mLSTM, with different word embeddings}

\subsection{Graph Generation Experiments}

In order to select the graph generation mechanism (c.f. Section~\ref{sec:graphbasedmodel}) that has the highest impact on the NLI problem, we ran experiments with match-LSTM as our text model and GconAttn as our graph model. GconAttn was chosen because the concepts retrieved from {\it One-Hop} and {\it Two-Hop} do not have any specific sequence or ordering. Table~\ref{tab:graph_gen_entailment_performance} presents the results of these experiments with the same hyperparameters for all three graph generation experimental conditions. 

All the graph+text models perform equally well. However, when only the graph model is considered, the {\it One-Hop} graph exhibits lower accuracy in comparison to the {\it Concepts Only} and {\it Two-Hop} graphs. This may be due to the noise induced from external knowledge and the addition of a large number of concepts in the {\it One-Hop} case. On average, premise and hypothesis sentences consist of $19$ and $12$ words respectively, but their respective {\it One-Hop} graphs have over $300$ concepts. On the other hand, the {\it Concepts Only} graphs average $9$ and $6$ concepts with better performance than the {\it One-Hop} graphs ($72.3\%$ vs $68.2\%$). Based on these results, and the simplicity of the graph construction technique, %in terms of the number of concepts and features input to our graph model, 
we pursued further experiments with the {\it Concepts Only} graphs.

The accuracy of the graph only models are relatively low, whereas the graph+text and text only models are comparable.  This might lead to the conclusion that the graph+text model is only driven by the text. However, an \textit{Oracle} model condition, where we choose the correct answer between the text and graph models, indicates that the graph model contributes to improved accuracy on the dev set (88.5\% for text only from Table~\ref{tab:entailment_performance} versus at least 91.6\% for Oracle). With \textit{One-Hop}, the graph model correctly predicts almost 40\% of the answers that are incorrectly predicted by the text-only model. We thus conclude that there is value to using external knowledge for NLI on SciTail. This is contrary to~\citet{MiClKhSa18}'s conclusion that external knowledge from ConceptNet is not useful in this domain.

\begin{table}[h!]
\resizebox{\linewidth}{!}{
\begin{tabular}{lccccc}  
\toprule
Graph & Graph & Graph + Text & Oracle & Avg Concepts & Avg Concepts \\
Generation & Model & Model & Text $\vee$ &Premise  &  Hypothesis \\
& Accuracy & Accuracy  & Graph & (19 words) & (12 words) \\
\midrule 
Concepts & 72.3	&  87.2 & \textbf{92.5} & 9 & 6 \\
One Hop & 68.2 & 87.3 & \textbf{93.1} & 369 & 312 \\
Two Hop  & 71.7 & 87.3 & \textbf{91.6} & 23 & 15\\
\bottomrule
\end{tabular}
}
\caption{Performance of graph generation techniques on the match-LSTM+GconAttn model on the SciTail dev set.}
\label{tab:graph_gen_entailment_performance}
\end{table}

\subsection{Selecting Text+Graph Model}
We experimented with many text models and selected match-LSTM due to its superior performance on SciTail. In order to determine the best combination of the graph models (Gmatch-LSTM and GconAttn) with match-LSTM as the text model, we ran multiple experiments on the SciTail dev set. The results are shown in Table~\ref{tab:combined_results}. Both GconAttn and Gmatch-LSTM are competitive in their performance. We also experimented with different knowledge graph embeddings to determine their impact on these models. ConceptNet-CN-PPMI clearly shows an improvement \textbf{(89.6)} in comparison to other models on the dev set. This model is used as our final model to be evaluated and compared against the baselines, leading to new state of the art performance numbers as shown in Table \ref{tab:entailment_performance}. 

\begin{table}[h!]
\resizebox{\linewidth}{!}{
\begin{tabular}{lllcc}  
\toprule
Text Model 	& Graph Model	&	Embedding					& Dev \\
\midrule 
%match-LSTM	&	--			&	--							& 88.5	\\
%AI2-Decomp	&	--			&	--							& 81.0	\\
%\midrule
match-LSTM	& Gmatch-LSTM	& ConceptNet-CN-PPMI  			& \textbf{89.6}	\\
match-LSTM	& Gmatch-LSTM	& ConceptNet-TransH 			& 87.3	\\
match-LSTM	& Gmatch-LSTM	& ConceptNet-CompleX 			& 88.3	\\
\midrule
match-LSTM	& GconAttn		& ConceptNet-CN-PPMI 			& 88.6	\\
match-LSTM	& GconAttn		& ConceptNet-TransH 			& 87.6	\\
match-LSTM	& GconAttn		& ConceptNet-CompleX 			& 88.4	\\
%\bottomrule
%AI2-Decomp	& Gmatch-LSTM	& ConceptNet-CN-PPMI 			& 75.2	\\
%AI2-Decomp	& Gmatch-LSTM	& ConceptNet-TransH 			& 78.8	\\
%AI2-Decomp	& Gmatch-LSTM	& ConceptNet-CompleX 			& 77.6	\\
%\midrule 
%AI2-Decomp	& GDecomp		& ConceptNet-CN-PPMI 			& 75.0	\\
%AI2-Decomp	& GDecomp		& ConceptNet-TransH 			& 78.8	\\
%AI2-Decomp	& GDecomp		& ConceptNet-CompleX 			& 75.5	\\
%\midrule 
%AI2-Decomp	& GDecompRNN	& ConceptNet-CCPPMI 			& 0.XX	& 0.XX \\
%AI2-Decomp	& GDecompRNN	& ConceptNet-TransH 			& 0.XX	& 0.XX \\
%AI2-Decomp	& GDecompRNN	& ConceptNet-CompleX 			& 0.XX	& 0.XX \\
%\bottomrule
%match-LSTM	& Gmatch-LSTM	& ConceptNet-CN-PPMI (length-ord) 			& \textbf{89.6}	& \textbf{85.2} \\
%match-LSTM	& Gmatch-LSTM	& ConceptNet-TransH (length-ord)			& 87.3	& 82.3 \\
%match-LSTM	& Gmatch-LSTM	& ConceptNet-CompleX (length-ord)			& 88.3	& 0.XX \\
%\midrule 
%match-LSTM	& GDecompRNN	& ConceptNet-CCPPMI 			& 0.XX	& 0.XX \\
%match-LSTM	& GDecompRNN	& ConceptNet-TransH 			& 0.XX	& 0.XX \\
%match-LSTM	& GDecompRNN	& ConceptNet-CompleX 			& 0.XX	& 0.XX \\
\bottomrule
\end{tabular}
}
\caption{Results of combinations of text and graph models along with various ways of computing embeddings on Scitail dev set.}

\label{tab:combined_results}
\end{table}

\section{Conclusion \& Future Work}\label{sec:conclusion_future_work}

In this paper, we presented the \nlisys~system: an entailment model for solving the Natural Language Inference (NLI) problem that utilizes ConceptNet as an external knowledge source. Our model provides performance that is close to state-of-the-art,  with an accuracy of 85.2\% on the SciTail dataset.
%Our model has close to state-of-the-art performance on the NLI classification problem, with an accuracy of $85.2\%$ on the SciTail dataset. 
% * <maknibassem@gmail.com> 2018-11-13T17:00:20.368Z:
% 
% > Our model has the current state-of-the-art performance on the NLI classification problem, with an accuracy of $85.2\%$ on the SciTail dataset. 
% We keep claiming that we have state of the art accuracy which is not the case. Are we ignoring HBMP and the leaderboard ? Or considering only the time the paper was written and not what happened in between ? 
% 
% ^.
%This translates to a improvement in performance on the ARC question answering dataset~\pavan{This is if we get a better performance using our model on ARC dataset -- RYAN todo}. 
We analyze various external knowledge sources and their effect on NLI, and show -- in direct contrast to other recent studies -- that there is promise in using knowledge graphs such as ConceptNet for textual entailment. 

% This is in direct contrast to other recent studies. 
% that have found that external knowledge does not improve performance on existing models for textual entailment. 
% Our results are also corroborated by an improvement in the accuracy of our model when knowledge sources are induced.  

Our future work includes designing a framework to exploit multiple relevant knowledge sources based on the given dataset and context. Existing external knowledge sources are known to be extremely noisy,  and new techniques must be developed in order to  extract knowledge relevant to a specific task (such as NLI). Another interesting direction involves exploring new ways to represent the structure of premise and hypothesis subgraphs, and systematically using the relations between the concepts contained therein to improve performance on the NLI task.

% Our work use only the concepts in the subgraphs generated for premise and hypothesis. The representations for the graphs are the graph embeddings for each concepts that is pre-trained. Exploring ways to represent the structure of the subgraphs for premise and hypothesis and methodically using the relations between concepts is an interesting area of research to improve the applicability of external knowledge and natural language inference. 

% {\small
\bibliographystyle{aaai}
\bibliography{abb,ureqa}
% }

\end{document}